# Text-to-hashtag Generation using Seq2Seq Learning

**Augusto Camargo[1], Wesley Carvalho[2], Felipe Peressim[3], Alan Barzilay[4], Marcelo Finger[5]**

[1]Department of Computer Science, Institute of Mathematics and Statistics

University of Sao Paulo, Brazil

R. do Matão, 1010 - Butantã, São Paulo - SP, 05508-090 - Brazil

augustoc@usp.br, wesley.seidel@gmail.com, felipe.peressim@ime.usp.br, alan.barzilay@usp.br, mfinger@ime.usp.br

***Abstract.*** *In this paper, we studied whether models based on BiLSTM and BERT can predict hashtags in Brazilian Portuguese for Ecommerce websites. Hashtags have a sizable financial impact on Ecommerce. We processed a corpus of Ecommerce reviews as inputs, and predicted hashtags as outputs. We evaluated the results using four quantitative metrics: NIST, BLEU, METEOR and a crowdsourced score. A word cloud was used as a qualitative metric. While all computer-generated metrics (NIST, BLEU and METEOR) indicated bad results, the crowdsourced results produced amazing scores. We concluded that the texts predicted by the neural networks are very promising for use as hashtags for products on Ecommerce websites. The code for this work is available at https://github.com/augustocamargo/text-to-hashtag.*

## 1. Introduction

This paper provides a method that addresses the problem of generating appropriate hashtags given the text in an Ecommerce review. There may be several reasons for the automation of such text generation, chief among which is enhancing the exposure of the content to search engines (SE).

Our main goal in this study was to generate good enough hashtags for SE attractiveness using neural networks. We decided to use BiLSTM and BERT in this paper because they achieve success in many NLP tasks [Devlin et al. 2018; Liu et al. 2019] and are currently used in a wide variety of tasks in Natural Language Processing (NLP).

We want to thank B2W Digital for providing the dataset used in this work. This work is possible thanks to their kindness["b2w-reviews01" [S.d.]]

## 2. Related work

Over the last 3 years, research on hashtag recommendations has become increasingly common [Li et al. 2016, 2019; Yang et al. 2019] [Kaviani and Rahmani 2020]. Our approach differs from other works primarily because we used a corpus of Ecommerce reviews instead of content from social media, we used classical and alternative metrics to quantify the results, and we used a novel approach to generate input for the BERT model.

On the Internet, there is an abundance of content regarding the implementation of neural networks for text translation and summarization similar to this work [Kaviani and Rahmani 2020; Li et al. 2016, 2019; Yang et al. 2019].

# 3. Methodology

We used a dataset containing Ecommerce reviews provided by B2W Digital. Our primary coding tool was JupyterLab ["Project Jupyter" [S.d.]]. The dataset was preprocessed before being used. The inputs to the LSTM and BERT were prepared specifically for each model. We describe both models in detail below.

## 3.1 Dataset

The language of the content in this dataset is Brazilian Portuguese. We used an open corpus of product reviews containing more than 132,373 Ecommerce customer reviews and 112,993 different user opinions regarding 48,001 unique products. The data were collected from the Americanas.com website between January and May 2018. We focused on two fields of the fourteen total present in the file:

> *review_title*: text format, introduces or summarizes the review content;
> *review_text*: text format, the main text content of the review.

## 3.2 Preprocessing

All digits and special characters were removed. We also padded punctuation with white spaces between punctuation and words.

### 3.2.1 BiLSTM

Two special tokens, '<start>' and '<end>', were added to review_title to help the decoder know where to start and end the decoding process. A sample from the processed corpus is:

> *review_title*: *<start> produto muito bom <end>*
> *review_text*: *excelente qualidade, chegou dentro do prazo, recomendo*

### 3.2.2 BERT

We used an autoregressive model for text generation over many iterations. Each iteration predicts an output word that is incorporated into the input of the next iteration. We predicted the output sentence from left to right, word by word. BERT was used in each iteration to predict the output word. We performed a classification task to refine BERT, increasing its ability to predict words. We want to emphasize that this model was adapted to predict words, as it is usually used for classification tasks.

A sample of the BERT-processed corpus is:

> fica super lindo ele aplicado , veio conforme o anunciado ! [SEP]  [MASK]
> fica super lindo ele aplicado , veio conforme o anunciado ! [SEP]  adorei
> fica super lindo ele aplicado , veio conforme o anunciado ! [SEP] adorei [MASK]
> fica super lindo ele aplicado , veio conforme o anunciado ! [SEP] adorei o
> fica super lindo ele aplicado , veio conforme o anunciado ! [SEP] adorei o [MASK]
> fica super lindo ele aplicado , veio conforme o anunciado ! [SEP] adorei o produto
> fica super lindo ele aplicado , veio conforme o anunciado ! [SEP] adorei o produto [MASK]
> fica super lindo ele aplicado , veio conforme o anunciado ! [SEP] adorei o produto !
> fica super lindo ele aplicado , veio conforme o anunciado ! [SEP] adorei o produto ! [MASK]
> fica super lindo ele aplicado , veio conforme o anunciado ! [SEP] adorei o produto ! [SEP]

### 3.3 Proposed Models

Several models were tested in preliminary experiments, and we describe the two that led to the best results: one for LSTM and one for BERT. All the details about the models can be seen here: https://github.com/augustocamargo/text-to-hashtag.

### 3.3.1 BiLSTM

Recurrent neural networks and their variations, LSTM and GRU, are a popular family of models in the NLP field specialized in modeling sequential data. We consider the bidirectional variants of this family of models. In this work, the bidirectional model is composed of LSTM layers (BiLSTM) with an attention mechanism between the encoder and decoder layers. The attention mechanism used is based on [Bahdanau et al. 2014].

The architecture of the BiLSTM used in this research was built as follows: both the encoder and decoder have an embedding layer at the input. The encoder has 3 LSTM layers with 32 cells each. The decoder has 2 LSTM layers with 64 cells each. The input and output dimensions were defined as 60 words per sentence and 15,997 words (the vocabulary size), respectively. The loss function chosen was sparse categorical cross entropy.

The teacher forcing technique was used to train the model more quickly. Using the teacher forcing method, we also pass the target data as input to the decoder. For example, when predicting 'hello', we pass 'hello' itself as an input to the decoder. This accelerates the training process. We use an inference model to predict our output sequences using the weights from a pre-trained model. In other words, the model generalizes what it has learned during the training process to handle new data.

### 3.3.2 BERT

BERT is a language model that stands for Bidirectional Encoder Representations from Transformers [Devlin et al. 2018]. This model is based on the core idea of [Radford et al. 2018]: apply the transfer learning technique from [Howard and Ruder 2018] using the transformer architecture. However, the model proposed by [Radford et al. 2018] is based on the transformer decoder, whereas BERT is based on the transformer encoder, which allows bidirectional predictions.

In short, BERT is a bidirectional transformer pre-trained using a combination of a masked language modeling objective and next sentence prediction on a large corpus comprising the Toronto Book Corpus and Wikipedia. According to [Devlin et al. 2018], BERT can be described as a stack of Lencoder layers, where the size of the vector output after passing through those layers is H and the number of self-attention heads in each layer is A. There are two versions of the BERT model: a simpler one, called BERT BASE, where $L = 12$, $H = 768$ and $A = 12$; and BERT LARGE, where $L = 24$, $H = 1024$ and $A = 16$.

In this research, the proposed model is based on the Huggingface library [Wolf et al. 2019] and was implemented with the help of the Keras API [Keras Team [S.d.]]. The reference model used was TFBertForMaskedLM. The input and output dimensions were defined as 72 words per sentence and 15,997 words (the vocabulary size), respectively. The loss function chosen was Sparse Categorical Cross Entropy.

### 3.4 Metrics

We quantitatively evaluate the results using four metrics: NIST, BLEU (Bilingual Evaluation Understudy), METEOR (Metric for Evaluation of Translation with Explicit ORdering) [Wołk and Koržinek 2016], and a crowdsourced metric.

We choose a crowdsourced [Wikipedia contributors 2020] score named metricF as a proxy to measure how attractive our hashtags are to a SE. The question for the annotator was, "Does this sentence look like a good or bad Ecommerce product page hashtag?". The scale ranges from 0 to 1: {0: meaningless or totally grammatically incorrect; 0.5: good context and average comprehensibility; 1.0: perfect context and completely comprehensible}.

A human annotator evaluated 6% of the total set of predicted sentences from both models (BiLSTM and BERT).

A word cloud was used as a qualitative metric to obtain an overview of the major words used in the texts, and its use allowed us to display text data in graphical form [DePaolo and Wilkinson 2014].

### 3.5 Experiments

We ran 17 experiments for the BiLSTM and 25 for BERT. Here, we present the best experiment for each model. All details regarding the other 16 experiments can be found at https://github.com/augustocamargo/text-to-hashtag.

In Table 1, all parameters used to execute the experiments for the BiLSTM and BERT are presented.

**Table 1: Setup of the experiments.**

|  | BiLSTM | BERT |
| --- | --- | --- |
| Training set | 81,746 | 81,746 (306,974) * |
| Validation set | 17,517 | 17,517 (66,014) * |
| Test set | 17,517 | 17,517 (65,657) * |
| Epochs | 16 of 20** | 5 of 5 |
| Batch | 128 | 128 |
| GPU | NVIDIA Tesla K80 - 12 Gb RAM | NVIDIA Tesla V100S - 32 Gb RAM |
| Execution time | 60 min | 225 min |

\* Size of the dataset after the preprocessing of the corpus for the BERT model.
\*\* The optimization goal was achieved before finishing all epochs.
**We used TensorFlow-GPU 2.3.1** ["TensorFlow" [S.d.]] **and Keras 2.4.3** [Keras Team [S.d.]]**.**

## 4. Results and Discussion

The implementation of the BiLSTM model was completely straightforward, but the BERT model was very counterintuitive and demanded many tests and debugging.

After more than 17 experiments, we believe that the BiLSTM model reached the limit of its performance for this corpus.

In Figures 1 and 2, we can see both models' performances during training. In Table 2, we present the optimization results: the BiLSTM model achieved good accuracy but the accuracy of BERT is not reliable because we used teacher forcing.

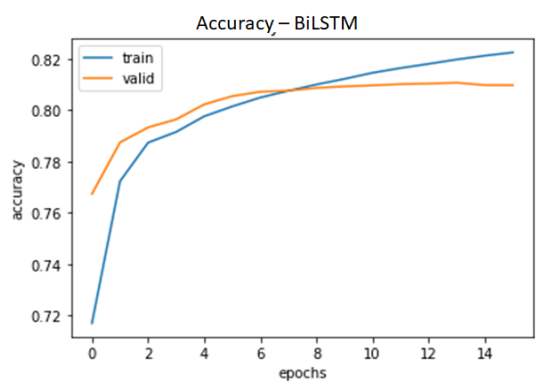

Figure 1: BiLSTM - training accuracy.

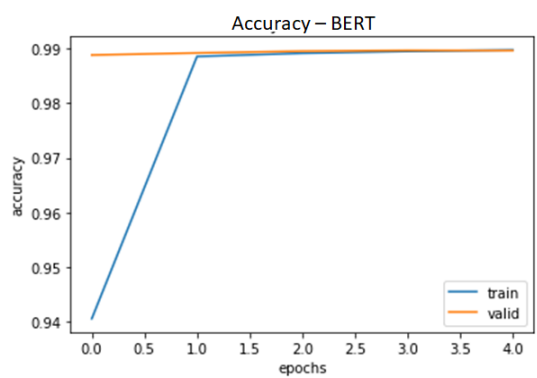

Figure 2: BERT - training accuracy.

In the BERT model, the accuracy validation was 0.990 and the test validation was 0.990. In the BiLSTM model, the accuracy validation was 0.809, and the test validation was 0.801. As seen in Figures 1 and 2, neither experiment overfits

The BERT model is 10x more *creative* than the BiLSTM: BERT created 1,491 unique sentences while the BiLSTM created just 135.

BERT is much more *literate* than the BiLSTM: BERT's vocabulary is almost 7x richer (relative to the original titles) than the BiLSTM. BERT used 21.37% (907 words) of the original vocabulary, and BiLSTM used just 3.29% (102 words)

The BiLSTM model is more *articulate* than BERT because it predicts sentences with 1-3 words, while BERT only predicts sentences with 1-2 words (Table 2). We can see that the BiLSTM model has a greater coefficient of variation (%CV) in the predicted title size (51.77%) than BERT (29.43%). Another insight from Table 2 is that the BiLSTM is *smarter* than BERT because it created larger sentences using a small vocabulary.

**Table 2: Size of words/sentences of original vs predicted texts.**

|  | BiLSTM - Average, SDev and %CV (words/sentence) | BERT - Average, SDev and %CV (words/sentence) |
| --- | --- | --- |
| Original text | 2.632 ± 1.647 %CV: 62.57% | 2.964 ± 1.866 %CV: 62.96 |
| Predicted text | 2.117 ± 1.096 %CV: 51.77 | 1.784 ± 0.525 %CV: 29.43 |

In Tables 3-4, we present samples of the predicted texts. In Tables 5-6 we can see scores for the predicted texts. NIST, BLEU and METEOR statistically calculated poor hashtag prediction results. However, metricF, our crowdsourced score, demonstrated particularly good results: 0.810 for the BiLSTM and 0.797 for BERT.

All computed metrics (NIST, BLEU and METEOR) produced bad results and did not address this task's needs. Computed metrics cannot capture semantics, while crowdsourced metrics (humans) can.

The BiLSTM metricF score obtained was 0.810 ± 0.349 and the BERT score was 0.912 ± 0.273. To have a baseline to metricF we asked the annotator to analyze the original text and the result was 0.959 ± 0.198. Using metricF and considering the standard deviation we could not distinguish the difference between the original and the generated hashtags.

Since computed metrics (NIST, BLEU and METEOR) are not on the same scale, we applied min-max normalization ["Max Normalization" [S.d.]] to the results:

**Table 3: BiLSTM - sample of predicted texts.**

| Review | Original | Predicted |
|---|---|---|
| atendeu todas minhas expectativas qualidade ótima entrega ótimo antes do prazo nota para o produto e para americanas | americanas melhor loja | excelente |
| isso é um descaso com o cliente já tem um mês que comprei e nunca chegou já até quebrei meu cartão americanas nunca mais compro aqui | nunca mais compro nessa loja | nao recebi o produto |
| a parte da frente que tem o unicórnio veio faltando a peça simplesmente não estava dentro da embalagem como pode ser resolvido esse problema absurdo o presente de natal da minha filha | veio peça faltando | não gostei do produto |

**Table 4: BERT - sample of predicted texts.**

| Review | Original | Predicted |
|---|---|---|
| comprei e não recebi o produto ! minha a avaliação vai para a americanas que não tem comprometimento com o cliente ! decepcionada ! | o produto não foi entregue | não recebi |
| jogo excelente , com gráficos ótimos e jogabilidade muito boa . recomendo a todos . | ótimo | excelente |
| ótimo custo benefício este aparelho atendeu perfeitamente minhas necessidades | aparelho bom | ótimo custo |

**Table 5: Scores - Values**

| | Score Average | |
|---|---|---|
| | BiLSTM | BERT |
| NIST | 0.066 ± 0.164 | 0.058 ± 0.140 |
| BLEU | 0.046 ± 0.121 | 0.058 ± 0.119 |
| METEOR | 0.107 ± 0.218 | 0.091 ± 0.188 |
| metricF | 0.794 ± 0.349 | 0.706 ± 0.389 |

**Table 6: Scores - % CV**

| | %CV of the Score | |
|---|---|---|
| | BiLSTM | BERT |
| NIST | 247.6% | 243.1% |
| BLEU | 263.9% | 205.5% |
| METEOR | 203.1% | 209.5%. |
| metricF | 43.8% | 55.1% |

The word clouds in Figures 3-5 demonstrate that words from predicted sentences for both the BiLSTM and BERT have very similar distributions. However, they are somewhat different from the Review words. This result suggests that the attention model truly works

**Word Cloud – Original x Predicted Sentences.**

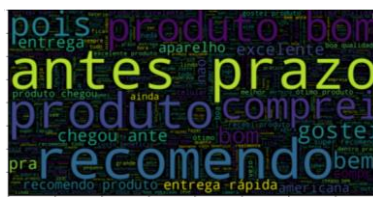 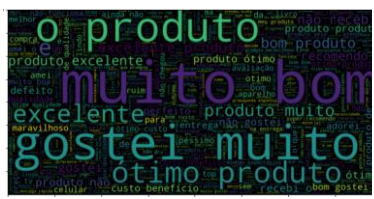 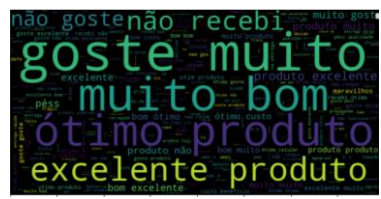

Figure 3: Original         Figure 4: BiLSTM         Figure 5: BERT

## 5. Conclusions

We achieved our main goal: hashtags predicted by the models are as attractive to SE as written by the humans; scores of metricF were almost the same for both computer-predicted and original sentences.

Our experimental results demonstrate that LSTM and transformers, when applied to text generation, were successful for this type of task. These models can therefore be used to predict Ecommerce hashtags using product reviews as inputs. We concluded that we could improve the chances of good SEO indexing using the hashtags generated by our neural networks and increase webpages traffic for Ecommerce coming from SEs. As said, with more traffic from SEO we have a lower cost of CAC and consequently a good impact on the Ecommerce profit.

Our future works will include different approaches and new architectures for generating sentences with BERT and new ways to measure SEO impact of hashtags.